\documentclass[a4paper,conference]{IEEEtran}

\usepackage{cite}

\usepackage{graphicx}
\usepackage{amsmath}
\usepackage{amssymb}
\newcommand{\R}{\mathbb{R}}

\usepackage{subcaption}                 % subfigure --> subfig --> subcaption
\usepackage[font=footnotesize]{caption}
\usepackage{tabularx,ragged2e}
\usepackage{multirow}
\usepackage{rotating}
\usepackage{colortbl}
\usepackage{arydshln}

\usepackage[capitalize]{cleveref}

\usepackage{fmtcount}									% english order (1st, 2nd, ...)
\usepackage{bm}                                         % some math command

\usepackage[printonlyused]{acronym}						% include acronym file (exclude from the list)
\acrodef {tcNet}[\textit{tc}Net]{Temporal Contextual Network}
\acrodef {MSE}[MSE]{Mean Squared Error}
\acrodef {BRNN}[BRNNs]{Bidirectional Recurrent Neural Network}
\acrodef {RNN}[RNNs]{Recurrent Neural Network}
\acrodef {CNN}[CNNs]{Convolutional Neural Network}
\acrodef {LSTM}[LSTM]{Long Short-Term Memory}
\acrodef {BLSTM}[BiLSTM]{Bidirectional Long Short-Term Memory}
\acrodef {MSE}[MSE]{Mean Squared Error}
\acrodef {NLP}[NLP]{Natural Language Processing}
\acrodef {AI}[AI]{Artificial Intelligence}
\acrodef {AE}[AE]{Autoencoder}

\usepackage[multiple]{footmisc}

\definecolor{color_dproposed}{RGB}{52,90,181}
\definecolor{color_battention}{RGB}{128,183,96}

\begin{document}


\title{Focusing on What is Relevant: Time-Series Learning and Understanding using Attention}
% Time series explainable using attention
% Others keywords : attention, sequencial, relation
% focusing, keyframe

\author{
	\IEEEauthorblockN{
Phongtharin Vinayavekhin{*},
Subhajit Chaudhury,
Asim Munawar,\\
Don Joven Agravante,
Giovanni De Magistris,
Daiki Kimura and
Ryuki Tachibana}
	\IEEEauthorblockA{IBM Research, Tokyo Japan\\
	%e-mail: \{pvmilk, subhajit, asim, ebb0spu, giovadem, daiki, ryuki\}@jp.ibm.com}
	*Corresponding author; E-mail: pvmilk@jp.ibm.com}
%TODO: change to corresponding email only
}

\maketitle

%%%%%%%%%%%%%%%%%%%%%%%%%%%%%%%%%%%%%%%%%%%%%%%%%%%%%%%%%%%%%%%%%%%%%%%%%%%%%%%%

\begin{abstract}

	This paper is a contribution towards interpretability of the deep learning
	models in different applications of time-series. We propose a temporal attention
	layer that is capable of selecting the relevant information to perform
	various tasks, including data completion, key-frame detection and
	classification. The method uses the whole input sequence to
	calculate an attention value for each time step. This results in more
	focused attention values and more plausible visualisation than previous
	methods. We apply the
	proposed method to three different tasks. Experimental results show that
	the proposed network produces comparable results to a state of the art. In
	addition, the network provides better interpretability of the decision,
	that is, it generates more significant attention weight to related frames
	compared to similar techniques attempted in the past.

\end{abstract}
\acresetall           				 % Reset all acronym definition in abstract

%%%%%%%%%%%%%%%%%%%%%%%%%%%%%%%%%%%%%%%%%%%%%%%%%%%%%%%%%%%%%%%%%%%%%%%%%%%%%%%%

%\begin{figure}[tb]
%    \centering
%	\includegraphics[draft, width=0.90\linewidth]{FIGURES/placeholder}
%    %\includegraphics[width=0.90\linewidth]{FIGURES/placeholder}
%    \caption{}
%    \label{fig:placeholder}
%\end{figure}

%%%%%%%%%%%%%%%%%%%%%%%%%%%%%%%%%%%%%%%%%%%%%%%%%%%%%%%%%%%%%%%%%%%%%%%%%%%%%%%%


\section{Introduction}

\label{section::introduction}

Recent progress in deep neural network has led to an exponential
increase in \ac{AI} applications. While most of these techniques have surpassed
human performance in many tasks, the effectiveness of these techniques and
their applications to real-world problems is limited by the
non-interpretability of their outcomes. Explainability is essential to
understand and trust \ac{AI} solutions.

Numerous techniques have been invented to gain insights
into deep learning models. These techniques provides post-hoc interpretability
to the learned model~\cite{conf:icml-ws:lipton2016}, which can be mainly
categorised into i) methods that perform a
calculation after training to find out what the model had learned without
affect the performance of the original model~\cite{conf:cvpr:mahendran15,
	conf:eccv:zeiler14}, and ii) model or layer that contains human
understandable information by construction~\cite{conf:nips:Alex12,
	conf:icml:xuc2015, conf:iclr:bahdanau15, conf:nips:vaswani2017}. This model
or layer generally improve or at least maintain model accuracy while providing
model insight. This paper follows the latter categories.

%
% Problem statement
%
In general, data can be characterised into two groups: spatial and temporal. In
this paper, we are interested in using a deep learning model to analyse
temporal data.
Learning structure of temporal data is crucial in many applications where the
explanation of the decision is as important as the prediction accuracy.
For instance, recognizing the most informative sequences of events and
visualizing them is useful and desired in computer
vision~\cite{conf:cvpr:tang12}, marketing analysis~\cite{conf:ms:markovitch08},
and medical applications~\cite{conf:nas:baker13}.
For example, if a patient were diagnosed with an illness, it is natural for
him/her to be curious on which information lead to this inference.

%
% How did we solve the problem
%
In this paper, we propose a novel neural network layer that learns temporal
relations in time-series while providing interpretable information about
the model by employing an attention mechanism. This layer calculates each
attention weight based on information of the whole time-series which allows
the network to directly select dependencies in the temporal data. Using the
proposed layer results in a focused distribution of the attention values and is
beneficial when interpreting a result of the network as it gives significant
weight to only the related frames.
This is in contrast to existing works for temporal
attention~\cite{conf:iclr:bahdanau15, conf:cvpr:pei2017} where the network
relies on \ac{RNN} to capture the temporal dependency of the input,
calculates each attention weight based on a single latent vector, and provides
more diffused attention which give significant weight to non-significant
frames.

We show how to use a proposed layer with a conventional neural network by
providing two architectures: auto-encoder and classification model.
These architectures are applied to three applications: motion capture data
completion, key-frame detection in video sequences, and action classification.
The experimental results show that the network achieves comparable accuracy to
state of the art and provides a clear focus on key frames that lead to the
outcome.

%%%%%%%%%%%%%%%%%%%%%%%%%%%%%%%%%%%%%%%%%%%%%%%%%%%%%%%%%%%%%%%%%%%%%%%%%%%%%%%%

\section{Related Works}
\label{section::related-work}

Deep learning models are often treated as black boxes.
However, it is important to understand what they are learning for certain applications.
Krizhevsky et al.~\cite{conf:nips:Alex12} show interpretability of the
network by visualizing weights of the first convolutional layer.
Mahendran and Vedaldi~\cite{conf:cvpr:mahendran15} try to understand what each
layer of a deep network is doing by inverting the latent representation using a
generic natural image prior.
Another approach is to interpret the function computed by each individual neuron.
This research can be separated into two categories: dataset-centric and network-centric.
The data-centric approach requires both the network and the data, while the latter approach only the trained network.
A dataset-centric approach displays a part of images that cause high or low activations for individual units.
Zeiler and Fergus~\cite{conf:eccv:zeiler14} propose a method that backtrack the network computations to identify which image patches are responsible for the activation of certain neurons.
Network-centric approach analyses the network without the availability of any data.
Nguyen et al.~\cite{conf:cvpr:nguyen15} use evolutionary algorithms or gradient descent to produce images that can fool neural networks.
Such techniques can be used to get a better understanding of the neural networks.

Instead of performing an additional calculation to visualise the model, this
paper focuses on one type of layer that contains interpretable information by
construction, an attention layer~\cite{conf:icml:xuc2015}.
This type of layer outputs
information, an attention matrix, that explains the network behavior.
Attention mechanism has become a
key component of sequence transduction for modeling the temporal dependencies.
Such mechanism is commonly used for temporal sequences together with \ac{RNN} and \ac{CNN}.
Bahdanau et al. \cite{conf:iclr:bahdanau15} provides an attention mechanism to
improve performance and visualisation of applications like machine
translation.
In this case, the attention is the value of the weights of a linear combination of a 
latent vector
encoded by \ac{RNN}.
Vaswani et al. \cite{conf:nips:vaswani2017} proposed a transformer architecture
for self-attention.
The transformer model relies entirely on self-attention to compute
representations of its input and output without using \ac{RNN} or \ac{CNN}.
M. Daniluk et al. \cite{conf:iclr:daniluk17} proposed a key-value attention
mechanism that uses specific output representations for querying a
sliding-window memory of previous token representations.
Sonderby et al.~\cite{conf:cln:SOnderby:2015} and Raffel and
Ellis~\cite{journals:corr:Raffel2015} modified the calculation of Bahdanau's
original attention mechanism~\cite{conf:iclr:bahdanau15}. Each of their
attention value is calculated as a function of the latent representation of a
\ac{BRNN} encoder of one current time step by assuming that the encoder can
capture the temporal information of the whole sequence.
In the proposed attention layer, each attention value depends on all time steps
of an input sequence. This allows the layer to compare
and choose the time steps that are more relevant to the desired output which
results in more focused attention value.

%%%%%%%%%%%%%%%%%%%%%%%%%%%%%%%%%%%%%%%%%%%%%%%%%%%%%%%%%%%%%%%%%%%%%%%%%%%%%%%%

\section{Proposed Method}

Here, we introduce the proposed neural network layer to learn
temporal relations of sequential data while allowing visualisation for model
interpretation. Next, we describe how to use the layer in two different network
architectures along with the details how to train them.

%Next, we describe how to use the layer in two different network architectures along with the details how to train them.
%Finally, the section explains how to train the network.

\subsection{Temporal Contextual Layer}
\label{section::tcl}

\begin{figure}[tb]
    \centering
    %\includegraphics[width=1.00\linewidth]{FIGURES/tcnet_allmodels}
	%trim={<left> <lower> <right> <upper>}
    \includegraphics[trim={0 0.6cm 0 0.6cm}, clip, width=0.98\linewidth]{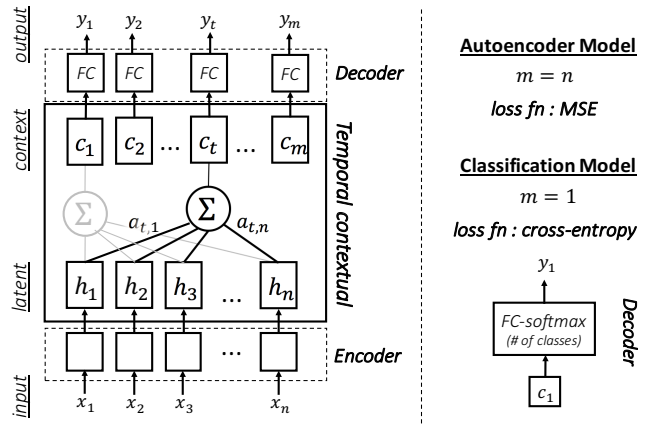}
	\caption{Temporal Contextual Layer: Neural network layer used to
		learn a temporal relation between two time-series. The layer takes an
		output of the time-distributed encoder as input, and calculated context
		vector as output. Types of architecture/model (auto-encoder or
		classification) depends on a time step of context vectors and the
		activation function of the decoder.}
    \label{fig:tcnet_allmodels}
\end{figure}

Our method assumes that some temporal relation exists in the time-series. The data is
not required to be precisely periodic only that there is some
semblance of temporal pattern. To learn the pattern, we propose a neural network layer which we refer to as a temporal contextual layer.
In addition, the layer has the advantage of interpretability as attention.
We begin by describing the proposed method as a layer that learn temporal
relation between two time-series in this section.

To start, we define input time-series as vector of length $n$ where each
vector element is a $g$-dimensional vector representing the current state. The
full input is a matrix: $\bm{H} = \begin{bmatrix} \bm{h}_1 &\bm{h}_2 &
	\dots &\bm{h}_{n} \end{bmatrix}^{\top}$ where the element at each time step
$t$ is $\bm{h}_t \in \R^{g}$. Similarly, the output sequence of length $m$ is
$\bm{C} = \begin{bmatrix} \bm{c}_1  &\bm{c}_2 & \dots &\bm{c}_{m}
\end{bmatrix}^{\top}$ where $\bm{c}_t \in \R^{g}$. Our layer is formulated
using an attention mechanism. Previously, an attention is used together
with \ac{RNN}~\cite{conf:iclr:bahdanau15} encoder and decoder on the
sequential data. It computes a context vector $c_t$ as the linear combination
of a sequence of~\ac{RNN} latent vector $h$:

\begin{equation}
	\bm{c}_t = \sum_{i=1}^{n}\alpha_{t, i}\bm{h}_i
	\qquad;\qquad
	\alpha_{t, i} = \frac{\exp{(e_{t, i})}}{\sum\limits_{i=1}^{n}\exp{(e_{t,i})}}
	\label{eq::one_step_lincomb}
\end{equation}
where $n$ is an input sequence length. $\alpha_{t, i}$ is a normalised weight
calculated by applying a softmax function on an attention weight $e_{t,i}$. An
attention weight is a learnable function $a$ of an input at a current time-step
$\bm{h}_i$ and a previous cell state $\bm{s}_{t-1}$ of the decoder $e_{t,i} =
a(\bm{s}_{t-1}, \bm{h}_i)$.

% TODO: careful with this statement, similar to intro
Although a latent representation of \ac{RNN} at one time-step $\bm{h}_i$ is a
function of all previous steps, it might not be able to capture long-term
information due its limited memory even with a gated-type like \ac{LSTM}.
Therefore, we propose to calculate an attention weight by providing an
information of all time steps to the layer:
\begin{equation}
	e_{t,i} = a(\bm{h}_1, \bm{h}_2, ..., \bm{h}_n) = a(\bm{H})
	\label{eq::attention_dependency}
\end{equation}

\cref{fig:tcnet_allmodels} shows temporal contextual layer.
It takes a predefined length sequence as input and
learns a mapping from this sequence to an output sequence, also of
predefined length (which may be different), using an attention mechanism.
Zero-padding can be used to handle time series with variable length.

In summary, two time-series $\bm{H}_{n\times g}$ and $\bm{C}_{m\times g}$ can be
temporally related by a matrix $\bm{A}_{m\times n}$ as:
\begin{equation}
	\bm{C} = \bm{A}\bm{H}	
	.
\label{eq::all_steps_lincomb}
\end{equation}
where $\bm{A} \in \R^{m\times n}$ is a temporal contextual matrix that
can be visualised in the same way as an attention matrix.
To achieve the desired behavior defined in~\cref{eq::attention_dependency}, it
can be defined as:
\begin{align}
	\bm{A} &=s_{r}(\bm{E})	\label{eq::all_steps_softmax}
	,\\
	\bm{E} &=\sigma_{v}(\sigma_{u}(\bm{U}\bm{H} + \bm{P})\bm{V} + \bm{Q})
	,
\label{eq::temporal contextual}
\end{align}
where $s_{r}$ is a row-wise softmax function and $\bm{E}\in \R^{m\times n}$ is 
the unnormalised temporal contextual matrix. $\bm{U}\in \R^{m\times n}$,
 $\bm{P}\in \R^{m\times g}$, $\bm{V}\in \R^{g\times n}$ and 
$\bm{Q}\in \R^{m\times n}$ are the learned weight and bias matrices and
$\sigma_{u}, \sigma_{u}$ are non-linear activation functions. 
Specifically, we use a tanh for $\sigma_{u}$ and a \textit{relu} for
$\sigma_{v}$. The proposed layer has a total of $2mn + gm+ gn$ of trainable parameters.

\subsection{Usages and Applications of Temporal Contextual Layer}
\label{section::usages_of_tcnet}

This subsection details two variations on how to combine the proposed layer with
conventional neural network layers to build a network architecture that can
solve a specific task.

\subsubsection{Autoencoder Model}
\label{subsection::autoencoder}

Temporal contextual layer is inserted between the encoder and the decoder, as
depicted in~\cref{fig:tcnet_allmodels}, to create an auto-encoder model that
learn temporal relations of time-series. Auto-encoder is an unsupervised
model that learns a representation of the data by generating an output to be
similar to the input it received.

%
% - explain the architecture.
%
Input time-series is encoded into latent representation
either by a dense layer or \ac{BRNN}. In the former, the layer is applied in
the timely-distributed manner, i.e. the same encoder is applied to each time step of
the input separately. The encoder could either be sparse or compressive. Then,
the time-series of the latent representation is passed to a temporal contextual
layer which outputs the time-series of the same length in time, $n=m$. Lastly, the
contextual latent representation is passed to the dense layer to decode the
time-series back to the same feature dimension as the original raw input.

An auto-encoder model can be used, for example, i) to perform data completion
and ii) to detect a key-frame in time-series.
For the former application, classically the denoising
auto-encoder model is well-known to be used on data
with random and partially occluded
data~\cite{conf:icml:vincent2008,conf:siggraph:holden2015}. The proposed
network can also be used for filling the occluded gaps in
time-series (data interpolation)~\cite{conf:nips:berglund2015}. This is due to its ability to find
the temporal relation in the time-series. \cref{subsection::h36m} demonstrates
this on motion capture data together with motion extrapolation task.
%Another application of the auto-encoder model is detecting a key-frame in a
%time-series.
For the latter application, the entire time-series is provided to the model as input, and the task is to
reconstruct only the desired key-frame as output. In this case, the proposed
layer learns to pick relevant information to reconstruct the desired key-frame.
This allows us to indirectly detect the key-frame from the attention weight
without explicitly training the network in a supervised fashion. Results are
shown in \cref{subsection::detecting_mnist} by detecting a key-frame in the
video.

\subsubsection{Classification Model}
\label{subsection::classification}

A temporal contextual layer can be used in a classification problem.
We consider one specific type of classification task
where the input is a sequential data and output is its corresponding class.
Examples of the real-world application are action recognition from a mo-cap
data, object recognition in video, speech recognition etc. Temporal contextual
layer of an output of $m=1$ time-step is placed before the final soft-max layer to
choose frames that are important to differentiate the time-series from others.
Similarly, a raw input sequence can either be encoded by a spatial layer or
\ac{BRNN} layer. The spatial layer can either be an encoder of one individual
frame or the encoder that combined information from multiple frames such as a 
convolutional network. The main idea here is to maintain the temporal order of the
input sequence. \cref{fig:tcnet_allmodels} shows a network architecture of
this classification model.

The proposed model provides insight on a result of the classification. Large
weights in an attention matrix specify the input frames that constitute to the
classification decision. We show this analysis in an action classification
experiment in~\cref{subsection::kit}.

%
% training the tcNet; the encoder and decode part can also be trained separately
%
Both auto-encoder and classification model with a temporal contextual layer can be treated as an
optimisation problem.
%like other deep learning models.
The loss function
is minimized through a stochastic gradient descent. The choice of the loss
function depends on whether the output time-series is discrete or continuous.
The gradient can be back-propagated through the layer as all operations in
the proposed layer are differentiable.

%%%%%%%%%%%%%%%%%%%%%%%%%%%%%%%%%%%%%%%%%%%%%%%%%%%%%%%%%%%%%%%%%%%%%%%%%%%%%%%%

\section{Experimental Results}
\label{section::experimental-results}

Temporal contextual network and the proposed architectures are applied to three
different tasks. First, we use an auto-encoder model to perform data completion
of mo-cap data and detect a key-frame in a video sequence. Finally, we use a
classification model to classify various human actions.

\subsection{Motion Capture Data Completion}
\label{subsection::h36m}

\begin{figure*}[tbh]
	\centering
    %\includegraphics[width=0.95\linewidth]{FIGURES/h36m_walking0_viz}
	%trim={<left> <lower> <right> <upper>}
    \includegraphics[trim={0 1cm 0 0.3cm},clip, width=0.95\linewidth]{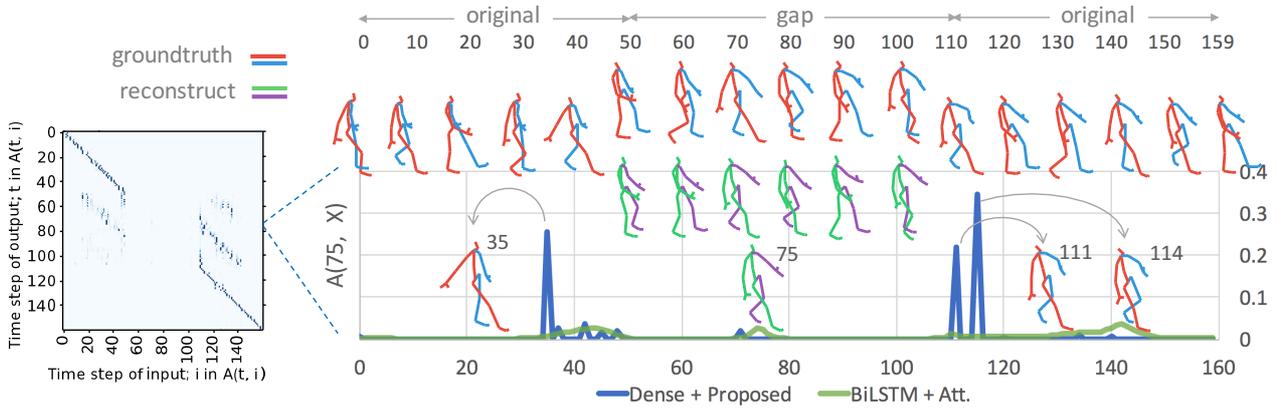}
	\caption{Result of motion capture completion (motion interpolation) of a walking sequence in a
		Human 3.6M dataset. Reconstructed poses are shown together with the
		original poses for a qualitative comparison. An attention matrix of all
		time steps, together with detailed attention weights of time step
		$t=75$ are
		shown in \textcolor{color_dproposed}{\rule[0.06cm]{0.4cm}{1.4pt}}. To
		reconstruct the pose of $t=75$, the network combines information from
		other poses with the similar appearance, $t=114, 111, 35$. Attention
		weights of previous method~\cite{conf:cln:SOnderby:2015} are also given
		in \textcolor{color_battention}{\rule[0.06cm]{0.4cm}{1.4pt}} for
		comparison.
	}
	\label{fig:h36m_walking0_viz}
\end{figure*}

%
% Explain the experiment and the dataset
%
For this task, we apply an auto-encoding model to fill the gap in occluded
motion capture data (motion interpolation) and to predict future motion (motion extrapolation).
A public dataset for 3D human motion, Human 3.6M~\cite{jour:pami:ionescu14}, is
used. The experiment is detailed as the following:
\begin{itemize}
\item The data is down-sampled to $25$ fps. Human posture is represented by
	joint orientation using an exponential map in the parent coordinate
	frame~\cite{conf:iccv:Fragkiadaki15,conv:cvpr:jian2016,conf:cvpr:martinez2017}.
\item For motion interpolation, each sequence is comprised of $160$
	frames. A zero-valued occluded hole of $60$ frames (2400ms) is created in
	the middle of the sequence; hence $50$ frames for both prefix and suffix
	motion. During training, these sequences are given as input while the
	output is the original sequences.
\item For motion extrapolation, 50 frames of prefix motion are used as input and
	the next 60 frames are used as output during training.
\item Motion of subject id $1,6,7,8,9,11$ is used for training and validation,
	while subject id $5$ is used for testing.
\item For each activity, there are $384$ training, $64$ validation and $8$
	testing sequences corresponding to the baseline for comparison
	purpose~\cite{conf:cvpr:martinez2017}. 
\item The auto-encoder model with a $2048$ neuron dense encoder is trained with
	a batch size of $8$ for $50$ epochs with \ac{MSE} loss.
\end{itemize}

%
% explain metric
% explain comparison methods
%
Results are evaluated using \ac{MSE} of joints in Euler angle, while the
difference in location and body rotation is
disregarded~\cite{conf:cvpr:martinez2017}. The results are
compared with convolutional
auto-encoders~\cite{conf:siggraph:holden2015} and other motion prediction
baselines~\cite{conf:iccv:Fragkiadaki15,conv:cvpr:jian2016,conf:cvpr:martinez2017}. 
Our implementation of~\cite{conf:siggraph:holden2015} follows the kernel size
reported, while the feature map of each layer is changed to $128, ~256, ~512$
respectively. Publicly available source code and
models~\footnote{https://github.com/asheshjain399/RNNexp}\footnote{https://github.com/una-dinosauria/human-motion-prediction}
are used to predict motion for the next 60 frames. We run all seq-to-seq
methods~\cite{conf:cvpr:martinez2017} and residual sampling-based loss (rSA)
method performs best in our experiment. Its average error of the last $100$
training iterations (out of a total of $10^{6}$) is reported in
\cref{table:h36m_result}.

\begin{table}[bt]
	%\tiny
	%\scriptsize
	\footnotesize
	%\small

	%160ms 320ms 560ms 1200ms 1840ms 2080ms 2240ms
	\begin{tabularx}{\linewidth}{l | X X X X X X X X}
		\hline
		\textit{Method}		    & \textit{160}   & \textit{320}   & \textit{560}   & \textit{1200}  & \textit{1840}  & \textit{2080}  & \textit{2240}  \\
								& \multicolumn{7}{c}{\textit{(milliseconds)}}					\\ \hline
\multicolumn{8}{c}{\textbf{Activity : Walking}}	\\ \hline
\scriptsize{CNN~\cite{conf:siggraph:holden2015}}
								& 1.31  & 1.35  & 1.33  & 1.44  & 1.45  & 1.51  & 1.44	\\
\scriptsize{\textbf{Proposed}}
								& \textbf{0.82}  & \textbf{0.82}  & \textbf{1.00}  & \textbf{0.75}  & \textbf{0.97}  & \textbf{1.02}  & \textbf{0.88} \\ \hdashline
\scriptsize{LSTM-3LR~\cite{conf:iccv:Fragkiadaki15}}
								& 0.98  & 1.37  & 1.78  & 2.24  & 2.40  & 2.28  & 2.39	\\
\scriptsize{ERD~\cite{conf:iccv:Fragkiadaki15}}
								& 1.11  & 1.38  & 1.79  & 2.27  & 2.41  & 2.39  & 2.49	\\
\scriptsize{S-RNN~\cite{conv:cvpr:jian2016}}
								& 0.94  & 1.16  & 1.61  & 2.09  & 2.32  & 2.35  & 2.36	\\

\scriptsize{S2S (rSA)~\cite{conf:cvpr:martinez2017}}
%								& \textbf{0.60}  & \textbf{0.87}  & 1.08  & 1.35  & 1.52  & 1.57  & 1.58 	\\
								& \textbf{0.59}  & \textbf{0.82}  & \textbf{0.97}  & 1.22  & 1.48  & 1.54  & 1.58 	\\ %% average values of final 100 results

\scriptsize{\textbf{Proposed}-pred}
 								& 1.11  & 1.16  & 1.04  & \textbf{0.99}  & \textbf{1.31}  & \textbf{1.35}  & \textbf{1.29}  \\ \hline

\multicolumn{8}{c}{\textbf{Actity : Smoking}}	\\ \hline
\scriptsize{CNN~\cite{conf:siggraph:holden2015}}
								& 2.02  & 2.24  & 2.31  & 2.45  & 2.49  & 2.37  & 2.17	\\
\scriptsize{\textbf{Proposed}}
								& \textbf{1.00}  & \textbf{1.19}  & \textbf{1.33}  & \textbf{1.43}  & \textbf{1.19}  & \textbf{1.04}  & \textbf{1.43} \\ \hdashline
\scriptsize{LSTM-3LR~\cite{conf:iccv:Fragkiadaki15}}
								& 1.61  & 1.98  & 2.27  & 2.52  & 2.60  & 2.76  & 2.93  \\
\scriptsize{ERD~\cite{conf:iccv:Fragkiadaki15}}
								& 1.80  & 2.19  & 2.54  & 3.45  & 3.36  & 3.33  & 3.36	\\
\scriptsize{S-RNN~\cite{conv:cvpr:jian2016}}
								& 0.94  & 1.16  & 1.61  & 2.09  & 2.32  & 2.35  & 2.36	\\
\scriptsize{S2S (rSA)~\cite{conf:cvpr:martinez2017}}
%								& \textbf{0.80}  & \textbf{1.21}  & \textbf{1.51}  & 2.11  & 2.38  & 2.47  & 2.51  \\
								& \textbf{0.76}  & \textbf{1.20}  & \textbf{1.46}  & 2.13  & 2.31  & 2.41  & 2.47  \\ %% average values of final 100 results
\scriptsize{\textbf{Proposed}-pred}
								& 1.09  & 1.27  & 1.52  & \textbf{1.91}  & \textbf{1.85}  & \textbf{1.96}  & \textbf{2.37} \\ \hline

\multicolumn{8}{c}{\textbf{Actity : Eating}}	\\ \hline
\scriptsize{CNN~\cite{conf:siggraph:holden2015}}
								& 1.33  & 1.44  & 1.55  & 1.90  & 1.62  & 1.62  & 1.39	\\
\scriptsize{\textbf{Proposed}}
								& \textbf{0.65}  & \textbf{0.85}  & \textbf{1.06}  & \textbf{1.39}  & \textbf{1.04}  & \textbf{1.01}  & \textbf{0.82} \\ \hdashline
\scriptsize{LSTM-3LR~\cite{conf:iccv:Fragkiadaki15}}
								& 1.25  & 1.82  & 2.28  & 2.69  & 2.65  & 2.74  & 2.70  \\
\scriptsize{ERD~\cite{conf:iccv:Fragkiadaki15}}
								& 1.79  & 2.33  & 2.61  & 2.42  & 2.35  & 2.37  & 2.30	\\
\scriptsize{S-RNN~\cite{conv:cvpr:jian2016}}
								& 1.41  & 1.85  & 2.19  & 2.84  & 2.99  & 3.05  & 3.11	\\
\scriptsize{S2S (rSA)~\cite{conf:cvpr:martinez2017}}
%								& \textbf{0.50}  & \textbf{0.80}  & \textbf{1.13}  & 1.72  & 1.71  & 1.85  & 1.87  \\
								& \textbf{0.50}  & \textbf{0.78}  & \textbf{1.10}  & 1.63  & 1.66  & 1.79  & 1.81  \\ %% average values of final 100 results
\scriptsize{\textbf{Proposed}-pred}
 								& 0.77  & 1.01  & 1.32  & \textbf{1.49}  & \textbf{1.58}  & \textbf{1.60}  & \textbf{1.51} \\ \hline

%\multicolumn{8}{c}{\textbf{Actity : Discussion}}\\ \hline
%\scriptsize{CNN~\cite{conf:siggraph:holden2015}}
%								& 2.44  & 2.52  & 2.59  & 2.44  & 2.71  & 2.64  & 2.48	\\
%\scriptsize{\textbf{Proposed}}
%								& \textbf{1.11}  & \textbf{1.26}  & \textbf{1.52}  & \textbf{1.60}  & \textbf{1.59}  & \textbf{1.68}  & \textbf{1.28} \\ \hdashline
%\scriptsize{LSTM-3LR~\cite{conf:iccv:Fragkiadaki15}}
%								& 2.12  & 2.25  & 2.33  & 2.48  & 2.50  & 2.49  & 2.48  \\
%\scriptsize{ERD~\cite{conf:iccv:Fragkiadaki15}}
%								& 2.47  & 2.68  & 2.92  & 3.23  & 3.23  & 3.44  & 3.47	\\
%\scriptsize{S-RNN~\cite{conv:cvpr:jian2016}}
%								& 1.49  & 1.83  & 2.07  & 2.24  & 2.53  & 2.60  & 2.56	\\
%\scriptsize{S2S (rSA)~\cite{conf:cvpr:martinez2017}}
%%								& \textbf{0.88}  & \textbf{1.22}  & \textbf{1.62}  & 1.86  & 2.06  & 2.13  & 2.03  \\
%								& \textbf{0.90}  & \textbf{1.22}  & \textbf{1.55}  & 1.99  & 2.24  & 2.26  & 2.18  \\ %% average values of final 100 results
%\scriptsize{\textbf{Proposed}-pred}
%								& 1.32  & 1.47  & 1.67  & \textbf{1.72}  & \textbf{1.68}  & \textbf{1.77}  & \textbf{1.71} \\ \hline
	\end{tabularx}
	\quad
	\caption{Comparing results of motion completion techniques.}
	\label{table:h36m_result}
\end{table}

%
% tested on four activities (walking, smoking, eating, discussion)
% show comparison table (cnn, lstm3lr, rnn, seq2seq).
%
%The proposed technique is applied to four activities, i.e. walking, smoking,
%eating and discussion.
Errors of three activities, i.e. walking, smoking, and eating, are reported
in~\cref{table:h36m_result}; above the dash line are results for motion
interpolation while extrapolation results are below. The proposed method
performs better than the other interpolation method (CNN). For extrapolation,
our method performs comparably in the short term against most methods, but
performs better in the long term. Interpolation results of the proposed method
also performs better than the extrapolation one. This is because the
interpolation method takes data before and after the gap to fill it.

%
% show the completion sequence, attention matrix, and analysis of the result (point to the frame)
%

\cref{fig:h36m_walking0_viz} displays reconstructed poses in various time steps
together with the original sequence. The attention matrix shows that the
network combines poses with similar appearance from different phases, both
before and after the time steps, to reconstruct a specific pose.
%It also learns not to take any information from an occluded gap.
This is more plausible than a previous attention method~\cite{conf:cln:SOnderby:2015}
(\cite{conf:iclr:bahdanau15} without \ac{RNN} decoder) that combines poses from
both similar and different appearances as depicted by a distributed attention
graph in the figure.
One disadvantage of the proposed method is that its output sequence is not smooth.
This is because the attention weight is not learned based on previous output.
One solution for this would be to use a \ac{RNN} decoder and incorporate the
knowledge of previously reconstructed frames into the attention
function~\cite{conf:iclr:bahdanau15}.

\subsection{Key-frame detection in a video sequence}
\label{subsection::detecting_mnist}

The proposed auto-encoder model is used for key-frame detection in this
experiment. We perform experiments to reconstruct MNIST
digits from a video sequence of randomly placed digits from $0$ to $9$. For
each video, there are 10 frames in total and each frame corresponds to a
MNIST digit. We specifically reconstruct the digit $2$ from the video input.
Convolutional encoder and decoder networks are pre-trained on MNIST images.
During training, we pass each image frame in the video through the encoder to
obtain feature space of size $(10, l)$, where $l=100$ is a latent dimension. These are passed
through the proposed layer which give an output of size $(1, l)$,
which is then passed through the decoder to reconstruct the desired image of
digit $2$. Encoder and decoder are also fine-tuned in this training process.

\begin{figure}[tb]
	\centering
	\begin{subfigure}{1.0\linewidth}
		\centering
		\includegraphics[width=0.90\linewidth]{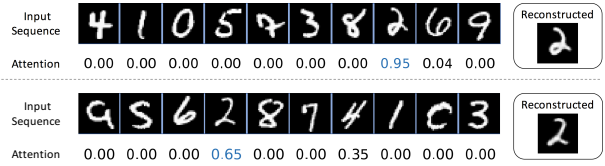}
		\caption{Samples of a good reconstruction and correct detection result.}
    	\label{fig:mnist_good_reconstructed}
	\end{subfigure}
	\begin{subfigure}{1.0\linewidth}
		\centering
		\includegraphics[width=0.90\linewidth]{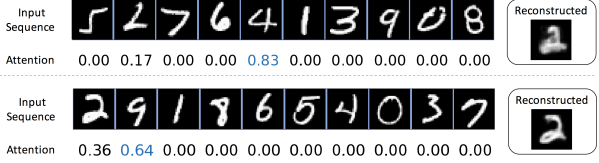}
		\caption{Samples of a bad reconstruction and incorrect detection result.}
    	\label{fig:mnist_bad_reconstructed}
	\end{subfigure}
	\caption{Attention value for each digit in the input video. For good
		reconstruction, the attention value on digit 2 is more compared to
		other digits.}
	\label{fig:mnist_keyframe}
\end{figure}

\cref{fig:mnist_keyframe} shows the attention value and reconstructed image.
Qualitative inspection shows that the proposed attention layer chooses
features from image frames containing digit $2$ if the reconstruction is of
good quality. In the failure cases, it obtains attention from other digits and
the reconstruction is poor. To quantitatively measure the performance of
proposed attention layer, we computed the detection accuracy of the location of
digit $2$ in the video sequence. For each video, we take the location of
maximum attention value at the detection of that input. We compared the
accuracy of this detection to the ground-truth location of digit $2$ in the
videos, yielding an accuracy of about $69\%$, which demonstrates that the
attention layer gives significant attention to pick features from those
locations. To encourage sparsity in the attention layer, we use a negative L2
activity regularisation for the attention unit, which acquires the minimum
possible value close to $-1$ when there is a single spike in the attention
activity.

\subsection{Action Classification on Motion Capture Data}
\label{subsection::kit}

% dataset
A proposed classification model is used to classify of human motion in the KIT
Whole-Body Human Motion Database~\cite{conf:icar:Mandery2015}. The motions are
captured using a VICON motion capture and fit to a human model to obtain a
sequence of joint angles. We select nine actions, listed
in~\cref{table:kit_classify_cm}, of two subjects (i.e. $3$ and $572$) based on
a balance of a number of data in each action. A total of $249$ motion sequences
are used. The experiment is detailed as the following:
\begin{itemize}
\item All sequences are down-sampled from $100$fps to $25$fps by
	selecting every forth frame. This increases the number of sequences
	by four times.
\item All sequences are padded with zeros to have the same length as the
	longest sequence (227 time steps). A posture in each time step contains 44
	joint angles; hence, each input sequence is a matrix of size $(277, 44)$.
\item A total of $996$ fps-reduced motion are divided into $697$
	training, $102$ validation and $197$ testing sequences.
\item The classification model with a $16$-neuron dense encoder is trained with
	a batch size of $17$ with a categorical cross-entropy loss. The training is
	stopped when validation loss decrease less than $0.01$ for at least $10$
	epochs.
\end{itemize}

%
% comparison of the classification, time/epoch to converged and number of parameters.
%
We compared the proposed method with a multilayer perceptron network (MLP) and
a multi-layer \ac{LSTM}~\cite{conf:cvpr:Shahroudy16}. In the former, we take
the proposed classification model with dense encoder and replace the temporal
contextual layer with a dense layer of $256$ neurons. In the latter, two layers
of \ac{LSTM} of $128$ cell units are concatenated and the output of the last
time steps is passed to a dense layer with softmax activation.

Results are evaluated using the accuracy and time required to
train the model. We ran an experiment $5$ times for each
method and the average are reported in~\cref{table:kit_acc_result}. The
classification model with dense encoder performs well in term of
accuracy. It takes less time to train than a multi-layer \ac{LSTM} because back
propagation through time is not required. While the method takes more time to
train comparing to MLP, it provides an interpretation of the result which will
be described later in the section. The proposed network also has a lower
number of parameters than other networks.

\begin{table}[tb]
	\centering
	\newcolumntype{Y}{>{\centering\arraybackslash}X}
	\begin{subtable}{1.0\linewidth}
		\begin{tabularx}{\linewidth}{l | Y  Y  Y}
			\hline
			\textit{Method} &
			\textit{Accuracy (\%)} &
			\textit{Training~Time~(s) \{Epochs\}} &
			\textit{Number of Parameters}\\
			\hline
			\scriptsize{MLP}
						& 76.3$\pm$1.1  & \textbf{13.4 \{12\}}  & 933,081   \\
			\scriptsize{Dense + \textbf{Proposed}}
						& 76.7$\pm$0.6  & 56.5 \{52\}  & \textbf{4,732}	\\
			\scriptsize{2 Layer LSTM~\cite{conf:cvpr:Shahroudy16}}
						& 65.7$\pm$7.7& 584.0 \{43\} & 221,321   				\\
			\scriptsize{BiLSTM + Att.~\cite{conf:cln:SOnderby:2015}}
						& 85.4$\pm$2.2 & 239.1 \{18\}  & 57,097	\\
			\scriptsize{BiLSTM + \textbf{Proposed}}
						& \textbf{85.9$\pm$2.7}  & 225.5 \{17\} & 86,252	\\ \hline
		\end{tabularx}
		\caption{Comparison of accuracy, training time, and number of parameters.}
		\label{table:kit_acc_result}
	\end{subtable}
	\begin{subtable}{1.0\linewidth}
		\begin{tabularx}{\linewidth}{c l | Y Y Y Y Y Y Y Y Y}
			\multirow{9}{*}{\begin{sideways}\textbf{True Label}\end{sideways}} &
			\textbf{B}ow 
				& 16 \cellcolor[gray]{0.79} & 0 & 0 & 0 & 0  & 0 & 0  & 0  & 0 \\
			& \textbf{J}ump 
				& 0 & 38 \cellcolor[gray]{0.5} & 0 & 0 & 0  & 0 & 0  & 0  & 0 \\
			& \textbf{K}ick 
				& 0 & 16 \cellcolor[gray]{0.79} & 16 \cellcolor[gray]{0.79} & 0 & 0  & 0 & 0  & 0  & 0 \\ 
			& \textbf{G}olf 
				& 0 & 0 & 0 & 17 \cellcolor[gray]{0.78} & 0  & 0 & 0  & 0  & 0 \\
			& \textbf{Te}nnis
				& 0 & 5 \cellcolor[gray]{0.93} & 0 & 0 & 22 \cellcolor[gray]{0.71} & 0 & 0  & 4 \cellcolor[gray]{0.95} & 0 \\
			& \textbf{Sq}uat 
				& 0 & 0 & 0 & 0 & 0  & 8 \cellcolor[gray]{0.89} & 0  & 0  & 0 \\
			& \textbf{St}omp 
				& 0 & 4 \cellcolor[gray]{0.95} & 0 & 0 & 0  & 0 & 12 \cellcolor[gray]{0.84}  & 0  & 0 \\
			& \textbf{Th}row
				& 0 & 4 \cellcolor[gray]{0.95} & 4 \cellcolor[gray]{0.95} & 0 & 0  & 0 & 0  & 8 \cellcolor[gray]{0.89} & 0 \\
			& \textbf{W}ave
				& 2 \cellcolor[gray]{0.97} & 2 \cellcolor[gray]{0.97} & 0 & 0 & 0  & 0 & 0  & 4 \cellcolor[gray]{0.95}  & 15 \cellcolor[gray]{0.80} \\ \cline{2-11}
		& & \textbf{B} & \textbf{J} & \textbf{K} & \textbf{G} & \textbf{Te} & \textbf{S} & \textbf{St} & \textbf{Th} & \textbf{W} \\
		 & \multicolumn{1}{c}{} & \multicolumn{9}{c}{\textbf{Predicted Label}} \\
		\end{tabularx}
		\caption{Confusion matrix of nine actions (\scriptsize{Dense + \textbf{Proposed}}).}
        \label{table:kit_classify_cm}
	\end{subtable}
	\caption{Result of action classification on a subset of KIT dataset.}
	\label{table:kit_results}
\end{table}

\begin{figure}[b]
    \centering
	\begin{subfigure}{1.0\linewidth}
        \centering
        \includegraphics[width=0.87\linewidth]{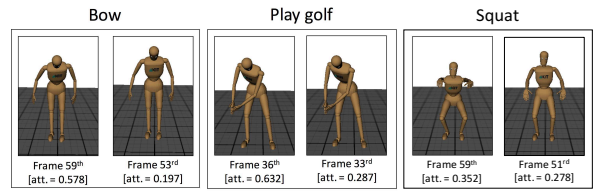}
		\caption{Postures with top-2 highest attention value (key frame) of
three actions with a perfect classification.}
        \label{fig:kit_good_postures}
	\end{subfigure}
	\hfill

	%set 1 : correct classification is similar, as well as the chosen frame.
	\begin{subfigure}{1.0\linewidth}
		\centering
		\includegraphics[width=0.87\linewidth]{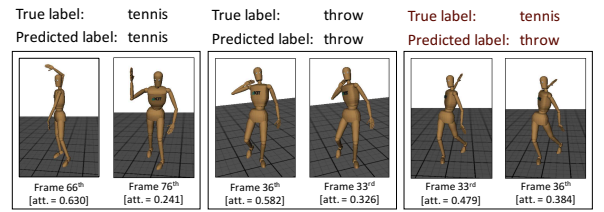}
		\caption{Classification where key frames of confusing classes are similar.}
		\label{fig:kit_postures_set1}
	\end{subfigure}
	\caption{Visualisation of result based on attention values
(\scriptsize{Dense + \textbf{Proposed}}).}
    \label{fig:kit_results}
\end{figure}

%
%confusion matrix and analysis of the visualisation result.
%

\cref{table:kit_classify_cm} shows a confusion matrix of the proposed
classification model. Three actions, i.e. \textit{bow}, \textit{play golf},
\textit{squat}, has a perfect classification result without any incorrect
classification to and from other actions. \cref{fig:kit_good_postures} shows a
sample of postures from those actions with the top-2 highest attention value
(key frames). Based on their attention value, information of these frames is 
used in the decision of the classification. These key frames are very different
between actions. They also correspond to human intuition to describe the
actions, but this is not always true for all actions as there is no control over
what the network will learn. On the other hand, a sample of postures with
incorrect classification results are shown in~\cref{fig:kit_postures_set1}. Key
frames for \textit{tennis} and \textit{throw} are similar which lead to some
incorrect classification.

%
%add result of bilstm+proposed method
%
\begin{figure}[tb]
	\centering
    \includegraphics[width=0.95\linewidth]{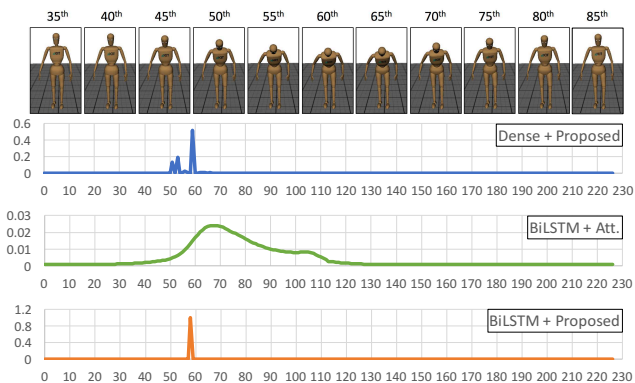}
	\caption{Comparison of attention value of various method in action
classification task. Proposed method provides more focus attention value to
classify \textit{bow} action (Note that Y-axis of all graphs are different).}
	\label{fig:kit_compare_attention}
\end{figure}

We compare the attention value with previous method by performing experiments
using~\ac{BLSTM} encoder together with the proposed attention layer
(BiLSTM + Proposed) and a feed-forward attention (BiLSTM +
Att.)~\cite{conf:cln:SOnderby:2015}. Methods with \ac{BLSTM} encoder perform
better in term of accuracy, but require more training time than the
dense encoder as shown in~\cref{table:kit_acc_result}.
A comparison of attention values of one of the \textit{bow} action is shown
in~\cref{fig:kit_compare_attention}. The action occurs between time step
\ordinalnum{50}-\ordinalnum{75}. The proposed attention layer with dense
encoder uses frames between the vicinity to make a decision, whereas the
\ac{BLSTM} encoder with a feed forward attention combines information both
inside and outside the action. Another downside when using \ac{BLSTM} encoder is
that one latent representation combines information from
various time steps, which makes it difficult to interpret the results. When
using the proposed layer with \ac{BLSTM} encoder, the
attention spike from very few frames, mainly frame \ordinalnum{58}.
In this case, the latent representation could have captured the temporal
information of the nearby frames that made it differentiable from other
actions.

%%%%%%%%%%%%%%%%%%%%%%%%%%%%%%%%%%%%%%%%%%%%%%%%%%%%%%%%%%%%%%%%%%%%%%%%%%%%%%%%

\section{Conclusion}
\label{section::conclusion}

%
% Conclusion
%
%The interpretability of a neural network has become a very important matter
%when using in applications that required an explanation of the result.
This paper proposes a neural network layer that learns the temporal
structure of the data.
The method is based on an attention mechanism which provides an interpretation to
the deep learning model.
We applied the method to various applications and showed
that using the proposed temporal contextual layer has retained, and in some
case improved, the performance of the model on the task.
The network also allows results to the model be interpreted and visualised.
As a future direction, we plan to investigate the idea to stack the temporal contextual layer to gain
insight into the deeper and more complex model.

%%%%%%%%%%%%%%%%%%%%%%%%%%%%%%%%%%%%%%%%%%%%%%%%%%%%%%%%%%%%%%%%%%%%%%%%%%%%%%%%


%%%%%%%%%%%%%%%%%%%%%%%%%%%%%%%%%%%%%%%%%%%%%%%%%%%%%%%%%%%%%%%%%%%%%%%%%%%%%%%%

%\section*{Acknowledgment}


%%%%%%%%%%%%%%%%%%%%%%%%%%%%%%%%%%%%%%%%%%%%%%%%%%%%%%%%%%%%%%%%%%%%%%%%%%%%%%%%



\bibliographystyle{IEEEtran}
\bibliography{IEEEabrv,reference}

\begin{thebibliography}{10}
\providecommand{\url}[1]{#1}
\csname url@samestyle\endcsname
\providecommand{\newblock}{\relax}
\providecommand{\bibinfo}[2]{#2}
\providecommand{\BIBentrySTDinterwordspacing}{\spaceskip=0pt\relax}
\providecommand{\BIBentryALTinterwordstretchfactor}{4}
\providecommand{\BIBentryALTinterwordspacing}{\spaceskip=\fontdimen2\font plus
\BIBentryALTinterwordstretchfactor\fontdimen3\font minus
  \fontdimen4\font\relax}
\providecommand{\BIBforeignlanguage}[2]{{%
\expandafter\ifx\csname l@#1\endcsname\relax
\typeout{** WARNING: IEEEtran.bst: No hyphenation pattern has been}%
\typeout{** loaded for the language `#1'. Using the pattern for}%
\typeout{** the default language instead.}%
\else
\language=\csname l@#1\endcsname
\fi
#2}}
\providecommand{\BIBdecl}{\relax}
\BIBdecl

\bibitem{conf:icml-ws:lipton2016}
Z.~C. Lipton, ``The mythos of model interpretability,'' \emph{{Int'l Conf. on
  Machine Learning (ICML), Human Interpretability in Machine Learning WHI}},
  2016.

\bibitem{conf:cvpr:mahendran15}
A.~Mahendran and A.~Vedaldi, ``Understanding deep image representations by
  inverting them,'' in \emph{Procs. of CVPR}, 2015.

\bibitem{conf:eccv:zeiler14}
M.~D. Zeiler and R.~Fergus, ``Visualizing and understanding convolutional
  networks,'' in \emph{Procs. of ECCV}, 2014.

\bibitem{conf:nips:Alex12}
A.~Krizhevsky, I.~Sutskever, and G.~E. Hinton, ``Imagenet classification with
  deep convolutional neural networks,'' in \emph{NIPS}, 2012.

\bibitem{conf:icml:xuc2015}
K.~Xu, J.~Ba, R.~Kiros, K.~Cho, A.~Courville, R.~Salakhudinov, R.~Zemel, and
  Y.~Bengio, ``Show, attend and tell: Neural image caption generation with
  visual attention,'' in \emph{Procs. of ICML}, 2015.

\bibitem{conf:iclr:bahdanau15}
D.~Bahdanau, K.~Cho, and Y.~Bengio, ``Neural machine translation by jointly
  learning to align and translate,'' in \emph{ICLR}, 2015.

\bibitem{conf:nips:vaswani2017}
A.~Vaswani, N.~Shazeer, N.~Parmar, J.~Uszkoreit, L.~Jones, A.~N. Gomez, L.~u.
  Kaiser, and I.~Polosukhin, ``Attention is all you need,'' in \emph{NIPS},
  2017.

\bibitem{conf:cvpr:tang12}
K.~Tang, D.~Koller, and L.~Fei-Fei, ``Learning latent temporal structure for
  complex event detection,'' in \emph{Procs. of CVPR}, 2012.

\bibitem{conf:ms:markovitch08}
D.~Markovitch and P.~N. Golder, in \emph{Using Stock Prices to Predict Market
  Events: Evidence on Sales Takeoff and Long-Term Firm Survival}, 2008,
  vol.~27, pp. 717--729.

\bibitem{conf:nas:baker13}
S.~J. Baker and E.~P. Reddy, ``Understanding the temporal sequence of genetic
  events that lead to prostate cancer progression and metastasis,'' \emph{Proc.
  of the National Academy of Sciences of the United States of America}, 2013.

\bibitem{conf:cvpr:pei2017}
W.~Pei, T.~Baltrusaitis, D.~M. Tax, and L.-P. Morency, ``Temporal
  attention-gated model for robust sequence classification,'' in \emph{In
  Procs. of CVPR}, 2017.

\bibitem{conf:cvpr:nguyen15}
A.~M. Nguyen, J.~Yosinski, and J.~Clune, ``Deep neural networks are easily
  fooled: High confidence predictions for unrecognizable images,'' in
  \emph{Procs. of CVPR}, 2015.

\bibitem{conf:iclr:daniluk17}
M.~Daniluk, T.~Rockt{\"{a}}schel, J.~Welbl, and S.~Riedel, ``Frustratingly
  short attention spans in neural language modeling,'' \emph{ICLR}, 2017.

\bibitem{conf:cln:SOnderby:2015}
S.~K. Sonderby, C.~K. Sonderby, H.~Nielsen, and O.~Winther, ``Convolutional
  lstm networks for subcellular localization of proteins,'' in \emph{Procs. of
  the Second Int'l Conf. on Algorithms for Computational Biology - Volume
  9199}, 2015, pp. 68--80.

\bibitem{journals:corr:Raffel2015}
C.~Raffel and D.~P.~W. Ellis, ``Feed-forward networks with attention can solve
  some long-term memory problems,'' \emph{Workshop Extended Abstracts of the
  4th ICLR, 2016}, 2015.

\bibitem{conf:icml:vincent2008}
P.~Vincent, H.~Larochelle, Y.~Bengio, and P.-A. Manzagol, ``Extracting and
  composing robust features with denoising autoencoders,'' in \emph{Procs. of
  ICML}, 2008.

\bibitem{conf:siggraph:holden2015}
D.~Holden, J.~Saito, T.~Komura, and T.~Joyce, ``Learning motion manifolds with
  convolutional autoencoders,'' in \emph{SIGGRAPH Asia 2015 Technical Briefs},
  ser. SA '15, 2015, pp. 18:1--18:4.

\bibitem{conf:nips:berglund2015}
M.~Berglund, T.~Raiko, M.~Honkala, L.~K\"{a}rkk\"{a}inen, A.~Vetek, and
  J.~Karhunen, ``Bidirectional recurrent neural networks as generative
  models,'' in \emph{NIPS}, 2015.

\bibitem{jour:pami:ionescu14}
C.~Ionescu, D.~Papava, V.~Olaru, and C.~Sminchisescu, ``Human3.6m: Large scale
  datasets and predictive methods for 3d human sensing in natural
  environments,'' \emph{{IEEE} Trans. Pattern Anal. Mach. Intell.}, vol.~36,
  no.~7, pp. 1325--1339, jul 2014.

\bibitem{conf:iccv:Fragkiadaki15}
K.~Fragkiadaki, S.~Levine, P.~Felsen, and J.~Malik, ``Recurrent network models
  for human dynamics,'' in \emph{Procs. of ICCV}, 2015.

\bibitem{conv:cvpr:jian2016}
A.~Jain, A.~R. Zamir, S.~Savarese, and A.~Saxena, ``Structural-rnn: Deep
  learning on spatio-temporal graphs,'' in \emph{Procs. of CVPR}, 2016.

\bibitem{conf:cvpr:martinez2017}
J.~Martinez, M.~J. Black, and J.~Romero, ``On human motion prediction using
  recurrent neural networks,'' in \emph{Procs. of CVPR}, 2017.

\bibitem{conf:icar:Mandery2015}
C.~Mandery, O.~Terlemez, M.~Do, N.~Vahrenkamp, and T.~Asfour, ``The kit
  whole-body human motion database,'' in \emph{Int'l Conf. on Advanced Robotics
  (ICAR)}, July 2015, pp. 329--336.

\bibitem{conf:cvpr:Shahroudy16}
A.~Shahroudy, J.~Liu, T.-T. Ng, and G.~Wang, ``Ntu rgb+d: A large scale dataset
  for 3d human activity analysis,'' in \emph{Procs. of CVPR}, 2016.

\end{thebibliography}

%%%%%%%%%%%%%%%%%%%%%%%%%%%%%%%%%%%%%%%%%%%%%%%%%%%%%%%%%%%%%%%%%%%%%%%%%%%%%%%%

%%%%%%%%%%%%%%%%%%%%%%%%%%%%%%%%%%%%%%%%%%%%%%%%%%%%%%%%%%%%%%%%%%%%%%%%%%%%%%%%

\end{document}